\documentclass[11pt,a4paper]{article}
\usepackage{arxiv}
\usepackage{times}
\usepackage{latexsym}
\usepackage[utf8]{inputenc}
\usepackage{multirow}
\usepackage{graphicx}
\usepackage{subcaption}
\usepackage{paralist}
\usepackage{color}
\usepackage{balance}

\usepackage{microtype}

\title{Multi-Round Parsing-based Multiword Rules for Scientific OpenIE}

\author{Joseph Kuebler, Lingbo Tong, Meng Jiang \\ University of Notre Dame}
\date{}

\begin{document}
\maketitle

\begin{abstract}
Information extraction (IE) in scientific literature has facilitated many down-stream tasks. OpenIE, which does not require any relation schema but identifies a relational phrase to describe the relationship between a subject and an object, is being a trending topic of IE in sciences. The subjects, objects, and relations are often multiword expressions, which brings challenges for methods to identify the boundaries of the expressions given very limited or even no training data. In this work, we present a set of rules for extracting structured information based on dependency parsing that can be applied to any scientific dataset requiring no expert's annotation. Results on novel datasets show the effectiveness of the proposed method. We discuss negative results as well.
\end{abstract}

\section{Introduction}

\begin{figure*}[h]
    \centering
    \includegraphics[width=0.9\linewidth]{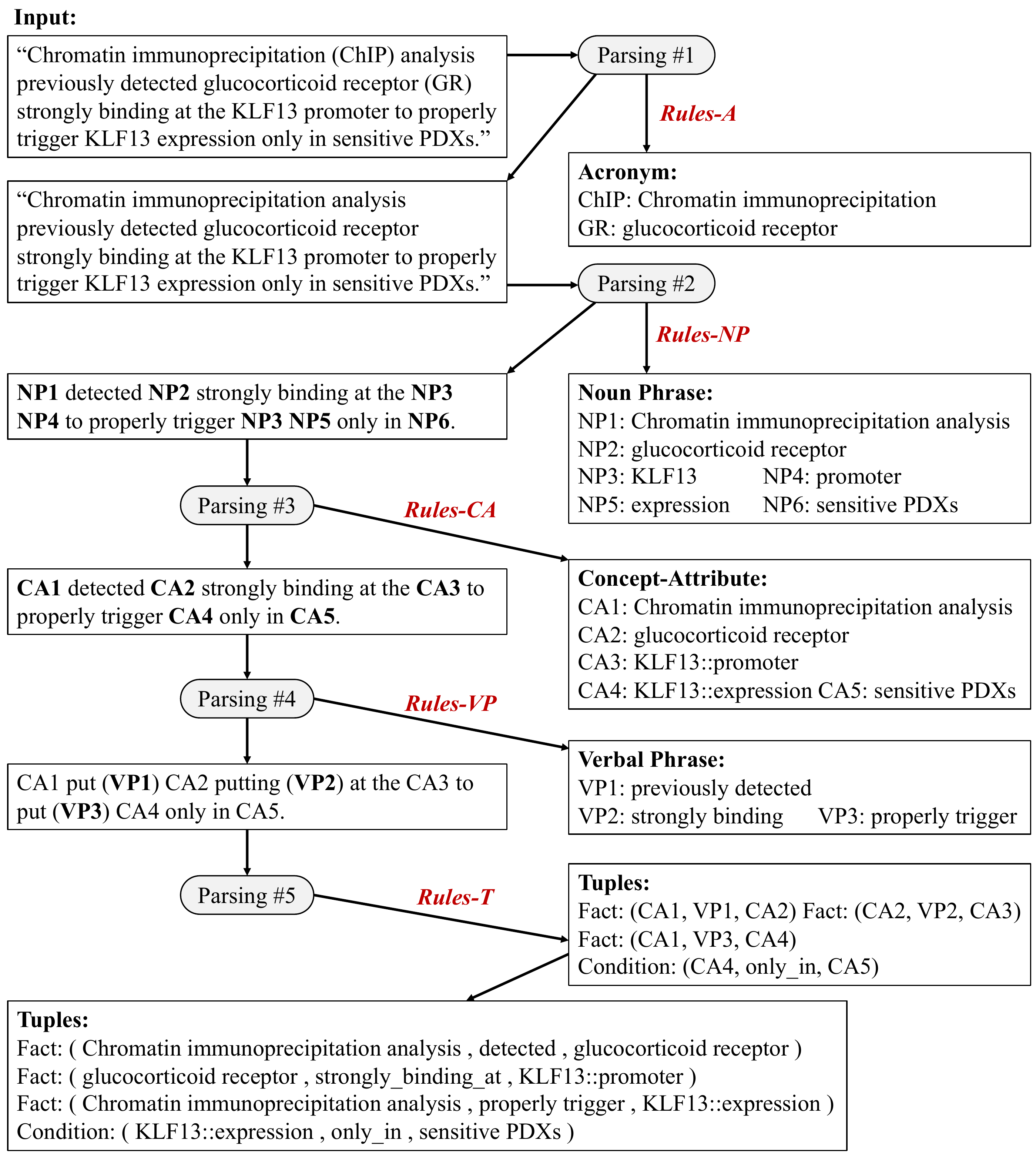}
    \caption{To turn scientific statement sentences (unstructured) into fact/condition tuples (structured), we perform multi-round parsing and design a set of rules to extract the structures step by step.}
    \label{fig:intro}
\end{figure*}

Mining structured information from scientific literature is important for many down-stream tasks such as document retrieval, knowledge discovery, and hypothesis generation.
In this work, we aim at this goal whose input and expected output can be illustrated by the example in Figure~\ref{fig:intro}.
The expectation has the following aspects, which leads the work to be \emph{unsupervised OpenIE} supported by a set of multi-round parsing-based multiword rules. Let's look at the bottom box in the figure
First, we argue that concept and relation recognition need to be \emph{open schema} because (a) no existing concept database (e.g., those in BioNLP, UMLS, MeSH) has all the concepts in these examples and (b) no existing relation schema has the relations such as ``detect'', ``bind'', and ``trigger''. So the target form of tuples has subject, object, and relational phrase such as ``strongly binding'' and ``only in''.
Second, the subjects and objects can be either concepts or \emph{attributes of concepts} such as promoter and expression of miRNA (e.g., ``KLF13::promoter'' and `KLF13::expression''), activation of specific pathways, and regulators of inhibitors.
Third, the tuples can have a role of either fact or \emph{condition of fact}. Conditions play an essential role in scientific statements: without the conditions that were precisely given by scientists, the facts might no longer be valid.
For example, the expression of KLF13 must be ``only in'' sensitive PDXs, which should not be removed or ignored in the output.

As shown in the figure, the process of extracting the tuples would have to identify multiple types of structured information such as acronyms, noun phrases (NPs, as concepts and attributes) and verbal phrases (VPs, as relations) that are often \emph{multiword expressions}. Acronym taggers and phrase chunkers are applied and studied: we find that (a) the extraction of one structure type can reduce the complexity of the sentence and (b) to accurately extract the multiwords, the dependency parsing tree can suggest ideas for the next round.

First, the dependency tags on the tree (e.g., ``compound'', ``appos'') help find acronyms even when the syntax is complex, like ``ChIP'' for ``\underline{ch}romatin \underline{i}mmuno\underline{p}recipitation''. Then the parenthesis are removed to simplify the sentence.

Second, the second round of parsing can help detect noun phrases. When the noun phrases are replaced by ``NP$x$'' ($x=1,2...$), the sentence can be significantly shortened, e.g., ``NP1 detected NP2 strongly binding at the NP3...''

Third, the third parsing identifies the relationship between the noun phrases: one NP is an attribute of the other NP, e.g, ``promoter of KLF13''; or a NP should be separated into two, one for concept and the other for attribute, e.g., ``KLF13 promoter''. Then we replace the concept or concept's attribute, which is a subject or object, by ``CA$x$'' ($x=1,2...$) so that all the remaining words in the sentence would not be related to the subjects or objects.

Fourth, another round of parsing can identify verbal phrases that might not be included in any existing relation schema. The phrases preserve adverbial modifiers that later can be used for learning and inference in down-stream tasks. The verbal phrases are tagged as ``VP$x$'' ($x=1,2...$).

Lastly, we perform parsing on the simplest variant of the sentence we can have. We design a set of rules to extract fact and condition tuples from the parsing tree. A lot of parsing and reorganizing the tree was done so that the final tree would be simpler, and patterns would be more common, when the initial trees could be quite large and the patterns were sometimes lengthy.

\textbf{Main contributions:} We build the multi-round parsing-based scientific OpenIE method and collected novel biomedical literature datasets, COVID-19 and miRNA for epithelial cancer. Each dataset has more than 50 sentences that were annotated by five experts in the corresponding domains. Our method performs much better than existing unsupervised OpenIE methods, however, we can still identify quite a few false predictions. We analyze the negative results by the end of in this paper.

\begin{figure*}[h]
    \centering
    \begin{subfigure}{0.48\textwidth}
      \centering
      \includegraphics[width=0.99\linewidth]{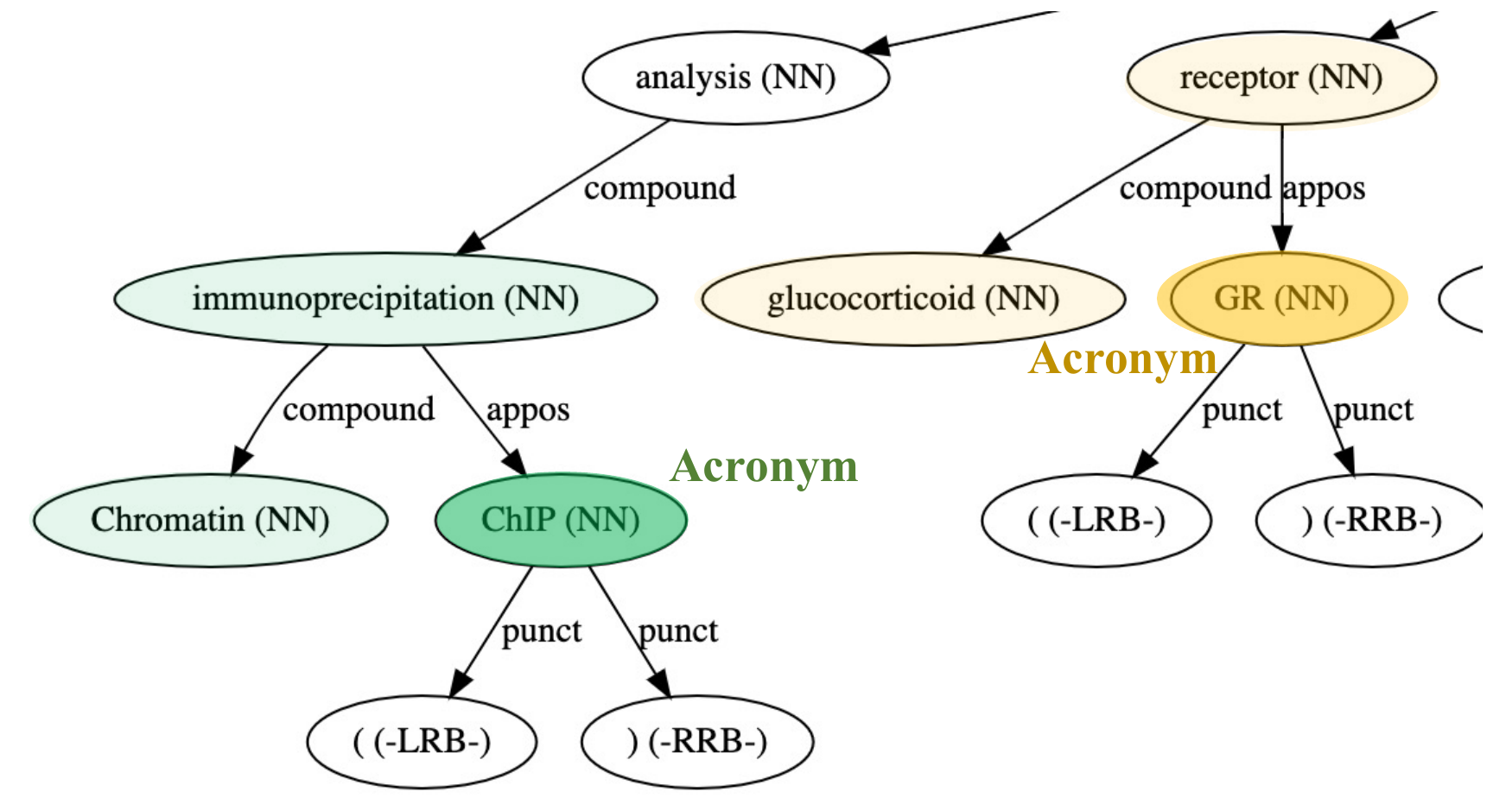}  
      \caption{Rules-A for \emph{acronym discovery} in \textbf{Parsing\#1}}
      \label{fig:parsing1}
    \end{subfigure}
    \hspace{0.1in}
    \begin{subfigure}{0.48\textwidth}
      \centering
      \includegraphics[width=0.99\linewidth]{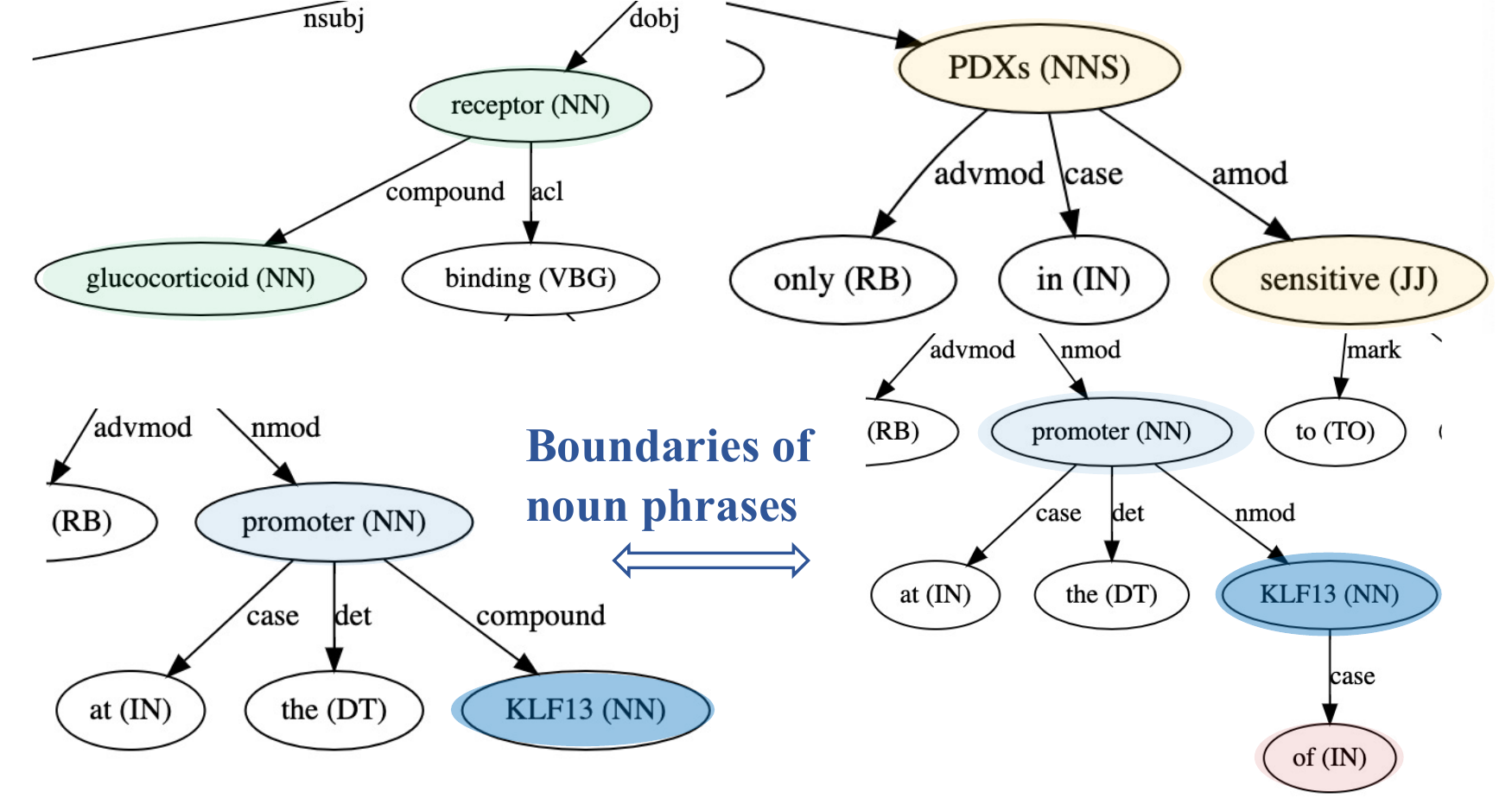}  
      \caption{Rules-NP for \emph{noun phrase chunking} in \textbf{Parsing\#2}}
      \label{fig:parsing2}
    \end{subfigure}
    \begin{subfigure}{0.48\textwidth}
      \centering
      \includegraphics[width=0.99\linewidth]{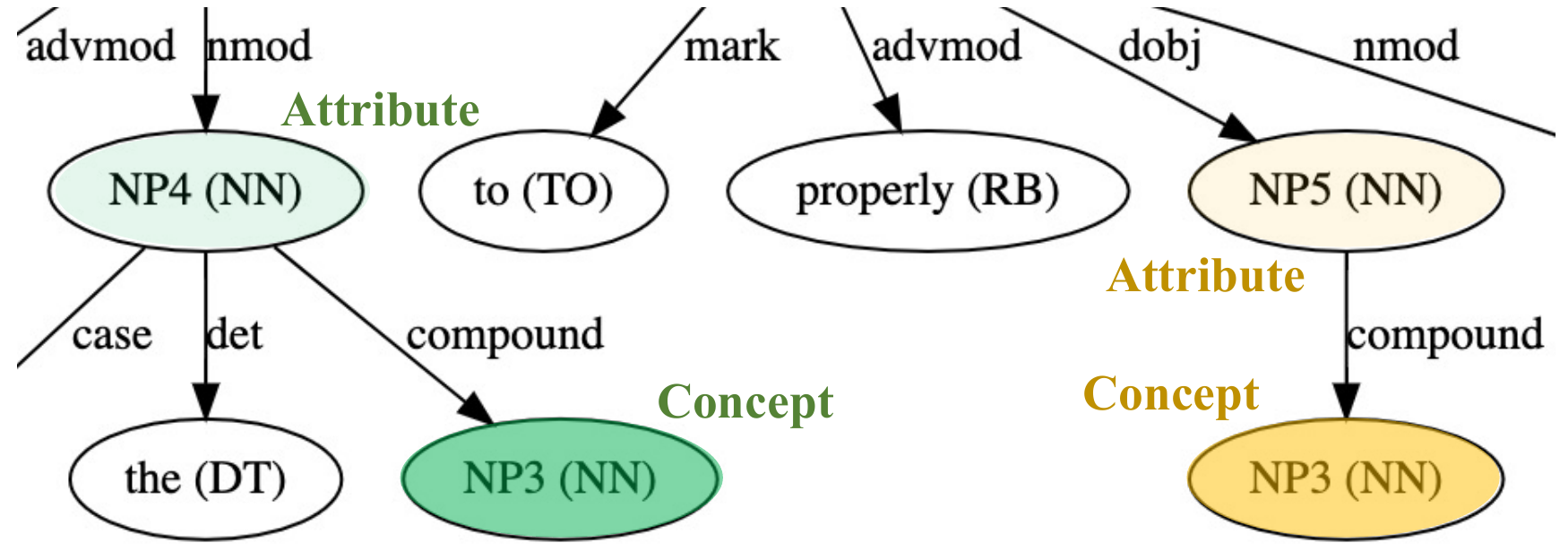}  
      \caption{Rules-CA for \emph{concept/attribute recognition} in \textbf{Parsing\#3}}
      \label{fig:parsing3}
    \end{subfigure}
    \hspace{0.1in}
    \begin{subfigure}{0.48\textwidth}
      \centering
      \includegraphics[width=0.99\linewidth]{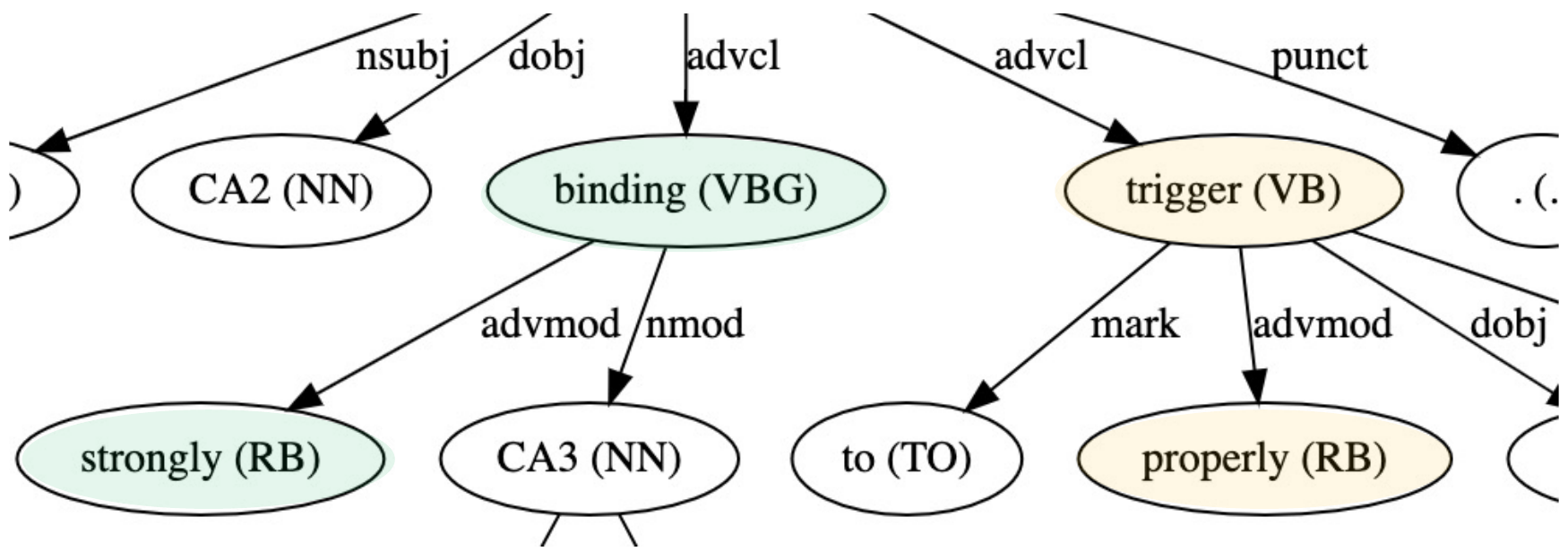}  
      \caption{Rules-VP for \emph{relational phrase chunking} in \textbf{Parsing\#4}}
      \label{fig:parsing4}
    \end{subfigure}
    \caption{Multiword rules for extracting preliminary structures (prior to tuples) in multi-round dependency parsing.}
    \label{fig:parsing1234}
\end{figure*}

\section{Related Work}

In this section, we review recent papers on three topics related with our work: scientific information extraction, OpenIE, and dependency parsing.


\subsection{Scientific Information Extraction}

Pre-trained scientific language models such as SciBERT \cite{beltagy2019scibert} and BioBERT \cite{lee2020biobert} have been widely used to perform specific information extraction tasks after being trained (called ``fine-tuned'') with a set of annotations on the tasks. \cite{park2020scientific} identified and classified scientific keyphrases with BERT models. \cite{yu2019biobert} detected named entities from electronic medical record with BioBERT. \cite{bai2020automatic} used BioBERT to detect health-related messages from Twitter data. \cite{jiang2020biomedical} detected BioBERT-based models to identify conditional statements and build biomedical knowledge graphs. There has not been much previously published work on unsupervised methods for scientific information extraction \cite{salloum2018using}.

\subsection{Open Information Extraction}

OpenIE techniques used to leverage linguistic structures such as dependency relations \cite{wu2010open,angeli2015leveraging}, clauses \cite{del2013clausie}, and numerical rules \cite{saha2017bootstrapping}. \cite{wang2018open} used meta textual patterns to discover open-schema facts in biomedical literature. \cite{stanovsky2018supervised} developed a supervised learning framework based on sequence labeling to leverage deep learning for OpenIE. \cite{muhammad2020open} used OpenIE to construct open-schema knowledge graphs. Conjunctive sentences \cite{saha2018open} and span models \cite{zhan2020span} were created for OpenIE, respectively. 
Recently, iterative learning methods become popular in OpenIE \cite{kolluru2020openie6,kolluru2020imojie}. Most existing scientific or biomedical IE methods use fixed relational schema to detect pair-wise relational facts. We argue the limitations in the introduction section and address them with OpenIE.

\subsection{Dependency Parsing}

A dependency parser analyzes the grammatical structure of a sentence, establishing relationships between ``head'' words and words which modify those heads \cite{rasooli2015yara}.
It generates a dependency parse tree that is a directed graph which has the following features: Root node can only be head in head-dependent pair. Nodes except Root should have only one parent/head. A unique path should exist between Root and each node in the tree \cite{gusfield1997algorithms}.
Recently, Allen Institute for AI developed SciSpaCy that contains fast and robust models for biomedical natural language processing \cite{neumann2019sciSpaCy}. It can support dependency parsing with three model options: small, medium, and large \cite{kanerva2020dependency}. The accuracy on part-of-speech tagging is as high as 0.9891. However, the F1 score on named entity recognition is 0.67--0.70, and thus, the performance on information extraction could even worse. We develop a pipeline to bridge the gap with multi-round parsing and multiword rules.

\section{Proposed Approach}

\begin{figure*}[t]
    \centering
    \includegraphics[width=1.\linewidth]{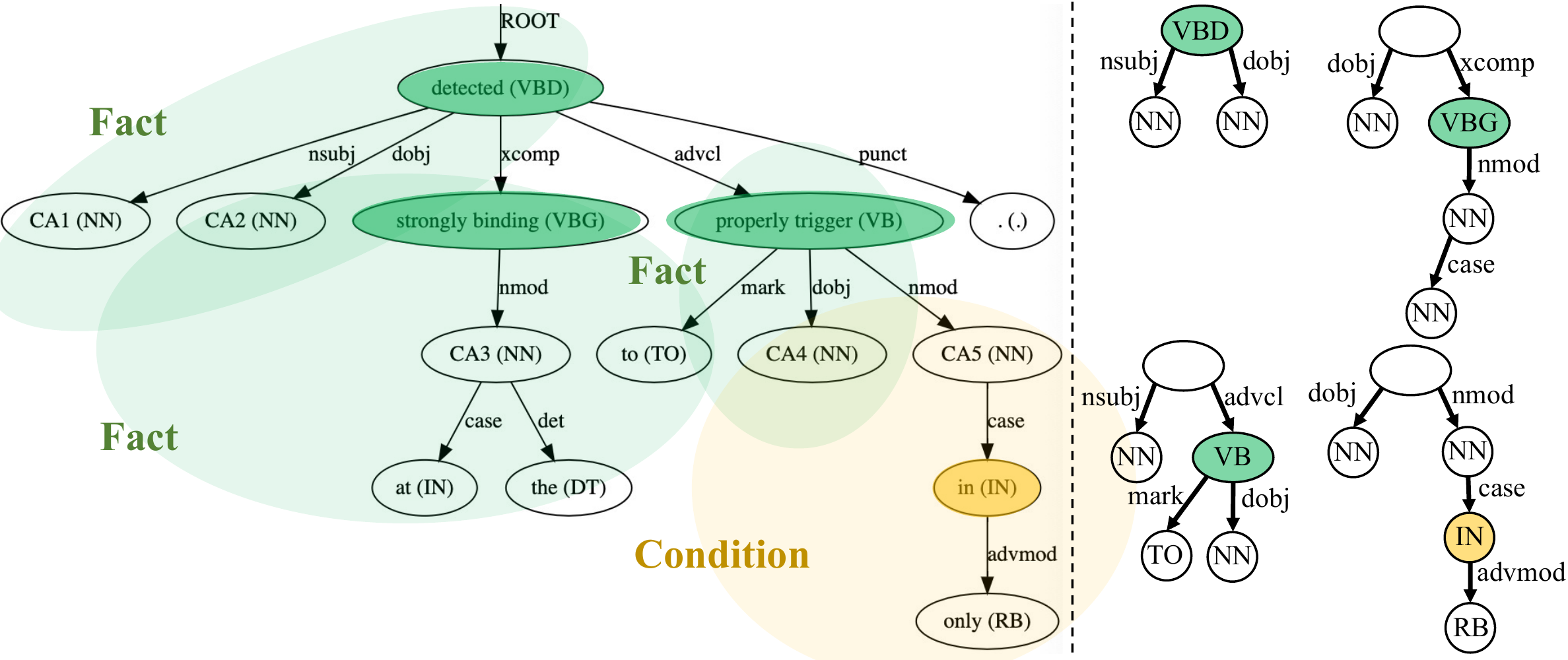}
    \caption{Rules-T for \emph{fact and condition tuple extraction} in \textbf{Parsing\#5}}
    \label{fig:parsing5}
\end{figure*}

In this section, we present our pipeline of \emph{five} rounds of dependency parsing and information extraction as well as the multiword rules for the tasks.

\subsection{Dependency parser} 

SciSpaCy is a Python package containing SpaCy models for processing biomedical or clinical text. The parsing model is named ``en\_core\_sci\_lg'' with a description as ``a full SpaCy pipeline for biomedical data with a larger vocabulary and 600k word vectors''. It creates a dependency parse tree where the nodes (i.e., tokens) have part-of-speech tag and the links between head-dependent pairs have dependency tag (see Figure~\ref{fig:parsing1234}).

\subsection{Acronym discovery: Rules-A in Parsing\#1} 

Acronyms can be identified based on the initials such as ``glucocorticoid receptor'' as ``GR''. However, sometimes the patterns could be complex like ``ChIP'' for ``chromatin immunoprecipitation''. The parse tree suggests the boundaries of the concept's name for identifying a new pattern (see Figure~\ref{fig:parsing1}).

\subsection{Noun phrase chunking: Rules-NP in Parsing\#2} 

Noun chunker is to group adjacent adjectives and nouns together to form noun phrases. They consisted of nouns and adjectives that together formed one phrase that could be referred to as a whole. This simplified the tree and reduced noise. For example, the adjective ``sensitive (JJ)'' can be put together with the noun ``PDXs (NNS)'' to form a phrase node ``sensitive PDXs (NP)'' to simplify the parse tree (see Figure~\ref{fig:parsing2}). The noun chunker leverages parsing patterns such as dependency ``amod'' between JJ and NNS, dependency ``compound'' between NN and NN, and numbers (labelled as CD) after NN (e.g., a nucleic acid stain ``Hoechst 33342''). One special case is that when the adjective is ``all'' or ``no'', it will be parsed as attribute instead of a modifier on a noun phrase. Another special case is that a pair of noun phrases could be expressed in multiple kinds of patterns such as ``A B'' (e.g., ``KLF13 promoter'') and ``B of A'' (e.g., ``promoter of KLF13''). Given two nouns ``A B'', they could be connected by tag ``compound''; only by leveraging ``B of A'' from sentences elsewhere, we could identify the separating boundaries of the noun phrases in ``A B''. 

\subsection{Concept and attribute recognition: Rules-CA in Parsing\#3} 

After parsing, we remove DT tokens (e.g., ``a'', ``an'', and ``the'') from noun phrases (NPs) in order to further simply the tree. Then we identify NPs connected by the preposition ``of (IN)''. This link typically indicates that one NP is an attribute of the other NP (see Figure~\ref{fig:parsing3}). The concept and its attribute are merged into one single node. The merged node and individual concept node are labelled as ``CA'' (for concept or attribute).

\subsection{Verbal (relational) phrase chunking: Rules-VP in Parsing\#4} 

A verb chunker is created to do similar simplification by grouping adverbs, helping verbs, and main verbs together to form a verbal phrase (VP) (see Figure~\ref{fig:parsing4}). Then an algorithm searches for a preposition connected to the verb under the ``prep'' dependency. If one is found, a noun phrase is searched for that is connected to the preposition under the ``pobj'' dependency. If one is found the preposition is attached at the end of VP. And if there is an extra adjective linked between the verb and the preposition, it will be put between the VP and preposition.

\begin{table*}[t]
\centering
\begin{tabular}{|l|r|r|r|r|}
\hline
\textbf{Dataset} & \textbf{\#Doc} & \textbf{\#Sent.} & \textbf{\#Facts} & \textbf{\#Cond.} \\ \hline \hline
TDMD-50 & 16 & 52 & 139 & 140 \\ \hline
COVID-200 & 21 & 216 & 349 & 355 \\ \hline
MDSC-600 & 25 & 612 & 945 & 1,164 \\ \hline
\end{tabular}
\caption{Three datasets annotated for Sci-OpenIE.}
\label{tab:datasets}
\end{table*}

\subsection{Fact and condition tuple extraction: Rules-T in Parsing\#5} 

An algorithm searches the parse tree for patterns or rules that could be used to form a tuple (see Figure~\ref{fig:parsing5}). The tuples can be classified into fact tuples and condition tuples. For fact tuples, the first rule or pattern looks for verbs that have a CA as a subject (``nsubj'' dependency) node as well as a CA as an object (``dobj'' or ``pobj'' dependency), forming (subject, verb, object)-tuples. Each time a tuple is formed, the function also searches for additional VPs or NPs that were connected to the tuple with the ``conj'' dependency. If a connection was found, a new tuple would form, replacing the original CA with the new one that was connected with the ``conj'' dependency.
The next rule iterates over all CAs in a sentence and searches for ones that have a VP node connected with the ``relcl'' dependency tag. It then checks if that verb node has a noun node connected with the object dependency tag.

For condition tuples, a rule looks for CAs that are connected through a preposition (IN). Each CA is iterated over and searching for preposition that linked to it. If a preposition was found, it is then iterated over to find another CA, forming (subject, preposition, object)-tuple. It should be noted that the preposition ``of'' is excluded because these has already parsed into noun phrases to form CAs.
Another pattern forms general condition tuples if there is no object found. This happens when a verb phrase is found, and then a CA is found connected with the subject dependency ``nsubj''. The next step is to find an object connected to the verb, but if none is found a general tuple would form with a ``NIL'' object, as (subject, verb, NIL).

\begin{table*}[t]
\centering
\begin{tabular}{|l|rr|rr|rr|}
\hline
& \multicolumn{2}{c|}{TDMD-50} & \multicolumn{2}{c|}{COVID-200} & \multicolumn{2}{c|}{MDSC-600} \\ \cline{2-7} 
& Micro F1 & Macro F1 & Micro F1 & Macro F1 & Micro F1 & Macro F1 \\ \hline \hline
IMoJIE & 0.034 & 0.066 & 0.033 & 0.067 & 0.026 & 0.051 \\ \hline
OpenIE6 & 0.040 & 0.076 & 0.045 & 0.085 & 0.029 & 0.055 \\ \hline
SOIE & 0.089 & 0.168 & 0.058 & 0.111 & 0.054 & 0.099 \\ \hline
\textbf{Ours} & \textbf{0.199} & \textbf{0.313} & \textbf{0.127} & \textbf{0.221} & \textbf{0.102} & \textbf{0.179} \\ \hline
\end{tabular}
\caption{Effectiveness of the proposed method: Compared with the OpenIE baseline methods, the proposed method achieves high Micro F1 and Macro F1 scores.}
\label{tab:baselines}
\end{table*}

\begin{table*}[t]
\centering
\begin{tabular}{|l|rr|rr|rr|}
\hline
& \multicolumn{2}{c|}{TDMD-50} & \multicolumn{2}{c|}{COVID-200} & \multicolumn{2}{c|}{MDSC-600} \\ \cline{2-7} 
& Micro F1 & Macro F1 & Micro F1 & Macro F1 & Micro F1 & Macro F1 \\ \hline \hline
Parsing\#1--\#5 & {0.199} & {0.313} & {0.127} & {0.221} & {0.102} & {0.179} \\ \hline
w/o Parsing\#2 (NP) & 0.186 & 0.292 & 0.112 & 0.183 & 0.099 & 0.172 \\ \hline
w/o Parsing\#4 (VP) & 0.161 & 0.255 & 0.108 & 0.182 & 0.090 & 0.156 \\ \hline
w/o Parsing\#5 (Tuple) & 0.162 & 0.262 & 0.110 & 0.187 & 0.090 & 0.161 \\ \hline
\end{tabular}
\caption{Ablation study: Compared with the variants of the proposed method, the multiple rounds of parsing and corresponding rules are effective.}
\label{tab:abalation}
\end{table*}

\section{Experiments}

In this section, we conduct experiments to demonstrate the effectiveness of our multi-round parsing-based method, compared with existing OpenIE methods. We will first present the dataset collection, baseline methods, and evaluation metrics, and then experimental results and our analysis.

\subsection{Datasets}

We recruited experts in the fields of target-directed miRNA degradation (TDMD) and myeloid-derived suppressor cell (MDSC) to annotate \emph{three} datasets. Data statistics are given in Table~\ref{tab:datasets}. About 16--25 documents were annotated for each dataset. The datasets are named according to the research domain of the literature and the number of annotated sentences. The smallest dataset TDMD-50 has only \emph{statement sentences} that were recognized by the experts. The medium-size COVID-200 was from the well-cited COVID-19 literature and only \emph{abstracts} were annotated. And the largest MDSC-600 annotated every sentence in the \emph{full-text}. The annotators follow a codebook about Scientific OpenIE to create fact and condition tuples. We observe that a sentence in abstract or full-text creates 1.5--1.6 facts and 1.6--1.9 conditions (see COVID-200 and MDSC-600); a statement sentence creates 2.7 facts and 2.7 conditions (see TDMD-50).


\subsection{Baseline Methods}

\noindent We compare with the following OpenIE methods:
\begin{itemize}
    \item SOIE \cite{angeli2015leveraging}: It was developed by a Stanford team and was one of the most popular OpenIE tools. It leveraged linguistic structures in dependency parse trees.
    \item OpenIE6 \cite{kolluru2020openie6}: It performed iterative grid labeling and coordination analysis to improve the performance of OpenIE.
    \item IMoJIE \cite{kolluru2020imojie}: It adopted an iterative memory-based framework that could produce the next extraction conditioned on all previously extracted tuples.
\end{itemize}
To compete with these methods in a fair environment, Parsing\#1 was performed to remove parentheses about acronyms in their input; and Parsing\#3 was performed to post-process their results for concept-attribute structures.

As we developed multiword rules for Parsing\#2 (noun phrase chunking), Parsing\#4 (verbal phrase chunking), and Parsing\#5 (tuple extraction), we will investigate the effect of each component in ablation studies by \emph{replacing} one parsing round and the corresponding rules by \emph{phrase chunkers and tuple extraction algorithms} in SciSpaCy and SOIE to create method variants.

\subsection{Evaluation Metrics}

For each method, we report final F1 scores using precision and recall \cite{martinez2018openie,lechelle2018revisiting,bhardwaj2019carb}. Precision represents the percentage of tuples that were correct and recall represents what percentage of possible tuples were found. This F1 was applied with a macro and micro method. Macro F1 treats each sentence as a separate instance and then averages all sentences scores together. Micro F1 treats all of the sentences as one big set and takes the F1 of everything as a whole. Both are useful. Micro accounts for certain sentences being more difficult than others and doesn't favor sentences with a small number of tuples or easier tuples. Macro gives a better idea of how effective the algorithm is on a random single sentence.

\subsection{Experimental Results}

We present and analyze the results in the experiments to answer the following questions: (1) Is the proposed method more effective than the OpenIE baseline methods? (2) Are the technical components, i.e., multi-round parsing and multiword rules, useful for the task of Scientific OpenIE?

\subsubsection{Effectiveness: Compared with OpenIE Baseline Methods}

Table~\ref{tab:baselines} presents the results on three datasets that compare our method with three baseline methods.

We observe that SOIE performs the best among the baseline methods though the other two methods are deep learning-based.
The main reason is that the rule-based SOIE can better be generalized to text from any scientific domains, while OpenIE6 and IMoJIE are not capable to extract the tuples accurately from the new scientific text -- they were trained on the annotated datasets in their original papers. The second reason that accounts for this observation is that SOIE can be easily adapted to SciSpaCy dependency parse tree by modifying the patterns used in Parsing\#1 and Parsing\#3 (concept-attribute structures).

\begin{figure*}[h]
    \centering
    \includegraphics[width=0.95\linewidth]{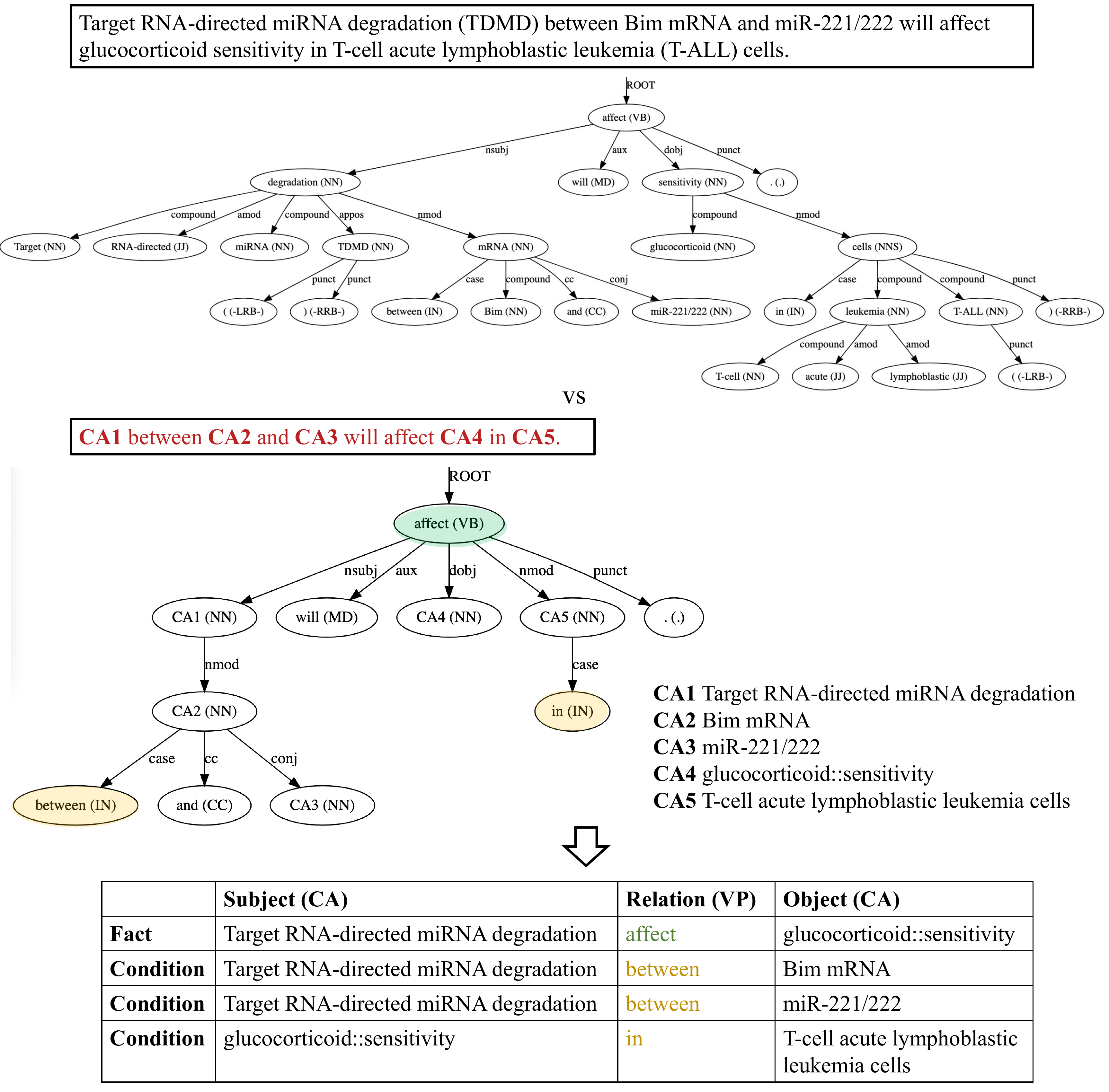}
    \caption{Case study: The multi-round parsing process can significantly simplify the structure of ``sentence''.}
    \label{fig:casestudy}
\end{figure*}

By comparing our method's results with the baselines', we observe that ours perform significantly better with higher Micro F1 and Macro F1 scores than SOIE. This is because our method captures more correct tuples by iterating through every subtree with a noun or verb as the root node in the dependency tree and matching them with designed patterns, while being able to exclude disguised wrong tuples with effective noun and verb phrase chunking strategies. In addition, our built-in steps for concept and attribute recognition account for its better performance for tuple extraction compared with traditional triple extraction methods.

We observe that the Micro F1 score is much lower than Macro F1 score. This is because the Micro F1 is affected heavily by the sentences that have many annotated tuples but very few extracted tuples, while the Macro F1 takes each sentence with equal weight and ignores the imbalance of tuple numbers. And we observe that the F1 score is lower when the dataset is bigger. This is because the larger dataset contains more various sentences, such as long sentences with multiple clauses and captions of figures and tables. The existence of various scientific expressions and special characters makes the tuple extraction more difficult.

\subsubsection{Ablation Study: Compared with Method Variants}

Table~\ref{tab:abalation} presents the results to compare the proposed method with the methods that replace particular components with conventional tools.

We observe that (a) the Micro F1 scores would be relatively 6.5\%, 11.8\%, and 2.9\% higher on TDMD-50, COVID-200, and MDSC-600, respectively; and (b) the Macro F1 scores would be relatively 6.7\%, 17.2\%, and 3.9\% higher, if we enabled Parsing\#2 and Rules-NP for noun phrase chunking. This is because the noun-chunking feature of the SpaCy library usually collected all the proper nouns, but did not always put them in the correct order. A common issue with the COVID data was that numbers following a word such as, ``coronavirus 2", would not always be grouped together. These functions fine-tuned the noun-chunking by reordering and editing the noun-phrases. Other examples included articles such as ``these" and ``no." ``These" was removed in instances where it began a noun-phrase, and no was changed from an adjective to an attribute (no\_symptoms $\rightarrow$ symptoms::no).

\begin{figure*}[t]
    \centering
    \includegraphics[width=\linewidth]{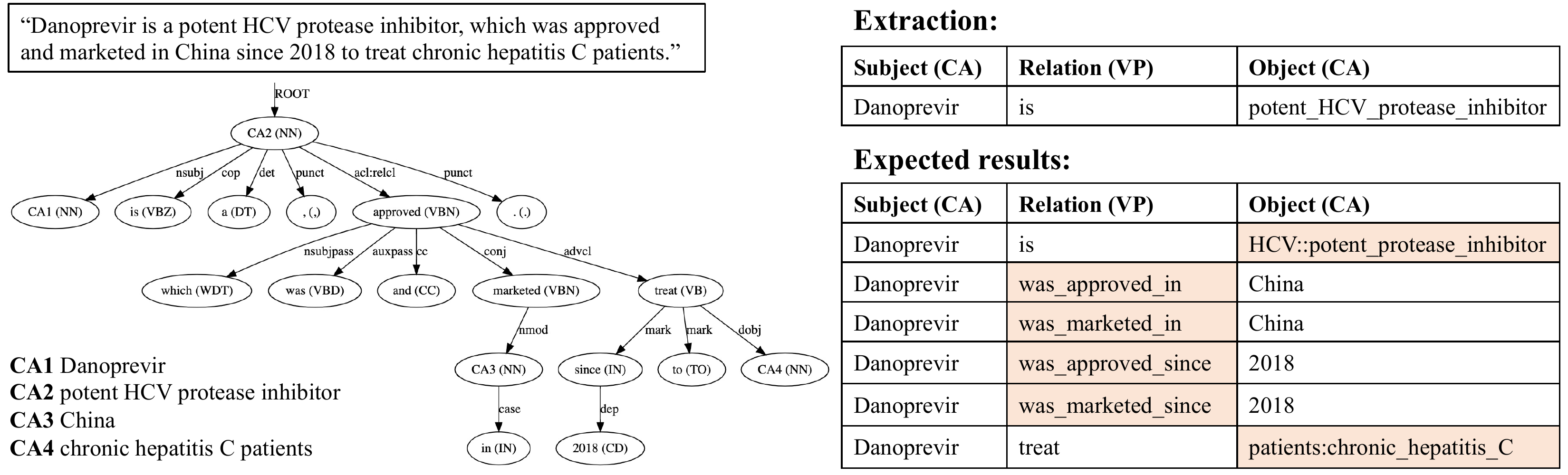}
    \caption{An negative example: The highlighted components in the expected results could not be extracted by the current design of extraction rules.}
    \label{fig:negative}
\end{figure*}

With Parsing\#4 and Rules-VP for verbal phrase chunking, we observe that (a) the Micro F1 scores would be relatively 19.1\%, 15.0\%, and 11.8\% higher and (b) the Macro F1 scores would be relatively 18.5\%, 17.6\%, and 12.8\% higher, on the three datasets respectively. This is because the verb-parsing function did not always pick up every word that should be included in the tuple. These functions and rules checked for the most common instances of the verb-phrase being incorrect and adjusted them. Most of the time, they were only missing one word that needed to be added to the phrase to make it correct.
In the example below:
\begin{quote}
    ``Cases quickly grew and spread throughout the hospital."
\end{quote}
the adverb ``quickly" should be added to both verbs (quickly\_spread\_throughout and quickly\_increased). Also, the preposition ``throughout" should be added to the VP for the first tuple. With the addition of these rules, both of these things now occur.

With Parsing\#5 and Rules-T for tuple extraction, we observe that (a) the Micro F1 scores would be relatively 18.6\%, 13.4\%, and 11.8\% higher and (b) the Macro F1 scores would be relatively 16.3\%, 15.4\%, and 10.1\% higher, on the three datasets respectively. This is because the tuple generation was adjusted to stricter rules. The NIL tuples were generated whenever a verb and and subject were found without an object. The tuple generation also better accounted for compound NPs and VPs. The rules were written to maximize precision. This is because rather than outputting all possible relations for NPs and VPs in a sentence, the phrases in a tuple had to all be directly related in a strict format.

\subsection{Case Study}

Figure~\ref{fig:casestudy} presents a case where the multi-round parsing process significantly simplified the structure of ``sentence''. Fact and condition tuples could be easily extracted from the simplest (final) parse tree. The tuples can later be used for knowledge graph construction, knowledge discovery, exploration, inference, and hypothesis generation.

\subsection{Discussion on Negative Results}

Figure~\ref{fig:negative} presents an instance of negative results. The first prominent issue was the depth at which each sentence was searched. As shown in the figure, the correct tuples formed a very complex pattern. In order to find such a pattern, the search depth must increase greatly. This was a very common occurrence, especially with complex sentences. Moving forward this could be expanded, but it would also require further optimization, because adding more depth to the search will increase the run time.

The second issue
is that the pre-processing did not correctly identify each CA and VP.
This is an issue with the SpaCy noun-chunking feature as well as the verb-chunking function. This was a pretty common issue in sentences that had multiple parentheses and numbers within it, such as this one. Moving forward, the processing and finding of CAs and VPs could be improved so that more would be recognized before converting the tree to tuples. 

\section{Conclusions}

In this work, we presented a novel pipeline that has multiple rounds of dependency parsing with a set of rules for extracting different types of structured information.
This pipeline was able to extract factual and conditional information from scientific literature in an OpenIE manner.
It did not require any relation schema or annotated data for model training.
The subjects, objects, and relations are often multiword expressions that need multiword rules for accurate extraction.
Results on novel datasets showed the effectiveness of this method. We discuss negative results as well.

\clearpage
\balance
\bibliographystyle{acl_natbib}
\bibliography{acl2021}


\end{document}